\documentclass[10pt,letterpaper,twocolumn]{article}
\usepackage[T1]{fontenc}
\usepackage[utf8]{inputenc}
\usepackage[margin=0.7in,columnsep=0.25in]{geometry}
\usepackage{newtxtext}
\usepackage{amsmath,amssymb,amsfonts}
\usepackage{graphicx}
\usepackage[authoryear,round]{natbib}
\usepackage{caption}
\usepackage{algorithm}
\usepackage{algorithmic}
\usepackage{booktabs,multirow}
\usepackage{xcolor}
\usepackage{placeins}
\usepackage{microtype}
\usepackage[hyphens]{url}
\usepackage[hidelinks]{hyperref}
\graphicspath{{figures/}}
\let\cite\citep
\let\shortcite\citeyearpar

\newcommand{\tableFullLocalization}{
\begin{table*}[t]
\centering
{\small
\setlength{\tabcolsep}{2.5pt}
\begin{tabular*}{\textwidth}{@{\extracolsep{\fill}}cccccccc@{}}
\toprule
\multirow{3}{*}{Method}
& \multirow{3}{*}{Venue}
& \multicolumn{6}{c}{Localization Recall ($\tau=5/10/15\,\mathrm{m}$) $\uparrow$} \\
\cmidrule(l){3-8}
& & \multicolumn{3}{c}{Validation Set} & \multicolumn{3}{c}{Test Set} \\
\cmidrule(lr){3-5}\cmidrule(l){6-8}
& & $k=1$ & $k=5$ & $k=10$ & $k=1$ & $k=5$ & $k=10$ \\
\midrule
Text2Pos~\shortcite{kolmet2022text2pos} & CVPR'22 & 0.14/0.25/0.31 & 0.36/0.55/0.61 & 0.48/0.68/0.74 & 0.13/0.21/0.25 & 0.33/0.48/0.52 & 0.43/0.61/0.65 \\
RET~\shortcite{wang2023ret} & AAAI'23 & 0.19/0.30/0.37 & 0.44/0.62/0.67 & 0.52/0.72/0.78 & 0.16/0.25/0.29 & 0.35/0.51/0.56 & 0.46/0.65/0.71 \\
Text2Loc~\shortcite{xia2024text2loc} & CVPR'24 & 0.37/0.57/0.63 & 0.68/0.85/0.87 & 0.77/0.91/0.93 & 0.33/0.48/0.52 & 0.61/0.75/0.78 & 0.71/0.84/0.86 \\
CMMLoc~\shortcite{xu2025cmmloc} & CVPR'25 & 0.44/0.62/0.68 & 0.75/0.88/0.90 & 0.83/0.93/0.95 & 0.39/0.53/0.56 & 0.67/0.80/0.82 & 0.77/0.87/0.89 \\
PMSH~\shortcite{feng2025pmsh} & ICCV'25 & 0.42/0.62/0.68 & 0.75/0.89/0.90 & 0.83/0.94/0.95 & 0.39/0.55/0.59 & 0.68/0.81/0.83 & 0.78/0.89/0.90 \\
MNCL~\shortcite{liu2025mncl} & AAAI'25 & \underline{0.54}/\underline{0.75}/0.79 & \underline{0.83}/0.94/0.95 & \underline{0.89}/\underline{0.97}/\underline{0.98} & \underline{0.50}/0.66/0.68 & \underline{0.76}/0.87/0.89 & \underline{0.83}/0.93/0.94 \\
Des4Pos~\shortcite{shang2026vehicle} & T-ITS'26 & 0.53/\underline{0.75}/\underline{0.80} & 0.81/\underline{0.95}/\underline{0.96} & 0.86/\underline{0.97}/\underline{0.98} & 0.48/\underline{0.67}/\underline{0.71} & 0.74/\underline{0.89}/\underline{0.90} & 0.81/\underline{0.94}/\underline{0.95} \\
\midrule
PARC-Loc & Ours & \textbf{0.71/0.85/0.87} & \textbf{0.91/0.97/0.98} & \textbf{0.94/0.98/0.99} & \textbf{0.67/0.80/0.82} & \textbf{0.87/0.94/0.95} & \textbf{0.90/0.96/0.97} \\
\bottomrule
\end{tabular*}
}
\caption{Full localization results on KITTI360Pose. Each entry reports Top-$k$ recall at $5/10/15\,\mathrm{m}$. Best results are in \textbf{bold}, and second-best results are \underline{underlined}.}
\label{tab:main_results}
\end{table*}
}

\newcommand{\tableCoarseRetrieval}{
\begin{table}[t]
\centering
{\small
\setlength{\tabcolsep}{2.2pt}
\begin{tabular*}{\columnwidth}{@{\extracolsep{\fill}}ccccccc@{}}
\toprule
\multirow{3}{*}{Method}
& \multicolumn{6}{c}{Submap Retrieval Recall $\uparrow$} \\
\cmidrule(l){2-7}
& \multicolumn{3}{c}{Validation Set} & \multicolumn{3}{c}{Test Set} \\
\cmidrule(lr){2-4}\cmidrule(l){5-7}
& $k=1$ & $k=3$ & $k=5$ & $k=1$ & $k=3$ & $k=5$ \\
\midrule
Text2Pos~\shortcite{kolmet2022text2pos} & 0.14 & 0.28 & 0.37 & 0.12 & 0.25 & 0.33 \\
Text2Loc~\shortcite{xia2024text2loc} & 0.32 & 0.56 & 0.67 & 0.28 & 0.49 & 0.58 \\
CMMLoc~\shortcite{xu2025cmmloc} & 0.35 & 0.61 & 0.73 & 0.32 & 0.53 & 0.63 \\
PMSH~\shortcite{feng2025pmsh} & 0.37 & 0.63 & 0.73 & 0.34 & 0.56 & 0.65 \\
MNCL~\shortcite{liu2025mncl} & \underline{0.50} & 0.75 & \underline{0.84} & 0.44 & 0.67 & 0.75 \\
Des4Pos~\shortcite{shang2026vehicle} & \underline{0.50} & \underline{0.76} & \underline{0.84} & \underline{0.45} & \underline{0.69} & \underline{0.77} \\
\midrule
PARC-Loc & \textbf{0.54} & \textbf{0.81} & \textbf{0.89} & \textbf{0.51} & \textbf{0.76} & \textbf{0.83} \\
\bottomrule
\end{tabular*}
}
\caption{Coarse submap retrieval recall on KITTI360Pose. Best results are in \textbf{bold}, and second-best results are
\underline{underlined}.}
\label{tab:coarse_retrieval}
\end{table}
}

\newcommand{\tableCoarseAblation}{
\begin{table}[t]
\centering
{\small
\setlength{\tabcolsep}{2.8pt}
\renewcommand{\arraystretch}{1.08}
\begin{tabular*}{\columnwidth}{@{\extracolsep{\fill}}ccccccc@{}}
\toprule
\multirow{2}{*}{Variant}
& \multicolumn{3}{c}{Validation Set}
& \multicolumn{3}{c}{Test Set} \\
\cmidrule(lr){2-4}\cmidrule(l){5-7}
& $k=1$ & $k=3$ & $k=5$ & $k=1$ & $k=3$ & $k=5$ \\
\midrule
Neural retriever & 0.49 & 0.75 & 0.83 & 0.46 & 0.70 & 0.78 \\
\addlinespace[2pt]
PARC w/o unary & 0.53 & 0.80 & 0.89 & 0.51 & 0.76 & 0.82 \\
PARC w/o relation & 0.52 & 0.79 & 0.87 & 0.50 & 0.74 & 0.82 \\
PARC w/o partial & 0.51 & 0.77 & 0.85 & 0.48 & 0.72 & 0.79 \\
\addlinespace[2pt]
\textbf{Full PARC} & \textbf{0.54} & \textbf{0.81} & \textbf{0.89}
& \textbf{0.51} & \textbf{0.76} & \textbf{0.83} \\
\bottomrule
\end{tabular*}
}
\caption{Retrieval-level ablation of PARC on KITTI360Pose. All variants share
our MNCL-style neural retriever. The full model is highlighted in
\textbf{bold}.}
\label{tab:coarse_ablation}
\end{table}
}

\newcommand{\tableFineAblation}{
\begin{table}[t]
\centering
{\small
\setlength{\tabcolsep}{4.0pt}
\renewcommand{\arraystretch}{1.10}
\begin{tabular*}{\columnwidth}{@{\extracolsep{\fill}}ccc@{}}
\toprule
\multirow{2}{*}{Cumulative variant}
& \multicolumn{2}{c}{Localization Recall $\uparrow$} \\
\cmidrule(l){2-3}
& Validation & Test \\
\midrule
Neural pipeline & 0.54/0.82/0.89 & 0.51/0.78/0.85 \\
$+$ Coarse PARC & 0.62/0.89/0.93 & 0.61/0.84/0.88 \\
$+$ Neighbor context & 0.66/0.90/0.93 & 0.64/0.85/0.89 \\
$+$ Query-conditioned priority & 0.69/0.90/0.93 & 0.66/0.85/0.90 \\
\textbf{Full PARC-Loc ($+$ bias)}
& \textbf{0.71/0.91/0.94} & \textbf{0.67/0.87/0.90} \\
\bottomrule
\end{tabular*}
}
\caption{Cumulative end-to-end ablation on KITTI360Pose. Each entry reports
Top-1/5/10 localization recall within $5\,\mathrm{m}$. The full model is
highlighted in \textbf{bold}.}
\label{tab:fine_ablation}
\end{table}
}

\newcommand{\tablePipelineDiagnosis}{
\begin{table}[t]
\centering
{\small
\setlength{\tabcolsep}{3pt}
\begin{tabular*}{\columnwidth}{@{\extracolsep{\fill}}cccc@{}}
\toprule
\multicolumn{4}{c}{\textbf{(a) Prevalence of Boundary-Truncated Evidence}} \\
\cmidrule(lr){1-4}
Analysis unit & Affected & Total & Rate \\
\midrule
Boundary-truncated instances & 7,907 & 19,629 & 40.28\% \\
Hints referencing them & 11,905 & 18,459 & 64.49\% \\
Queries & 3,179 & 3,187 & 99.75\% \\
\bottomrule
\end{tabular*}

\vspace{5pt}

\begin{tabular*}{\columnwidth}{@{\extracolsep{\fill}}ccccc@{}}
\toprule
\multicolumn{5}{c}{\textbf{(b) Submap Retrieval Outcomes}} \\
\cmidrule(lr){1-5}
& \multicolumn{2}{c}{Top-1 Recall}
& \multicolumn{2}{c}{Top-1 Retrieval Errors} \\
\cmidrule(lr){2-3}\cmidrule(l){4-5}
Model & Top-1 & Close@1 & Close non-GT & Other \\
\midrule
Text2Loc~\shortcite{xia2024text2loc} & 33.0\% & 64.3\% & 46.8\% & 53.2\% \\
CMMLoc~\shortcite{xu2025cmmloc}  & 34.7\% & 66.3\% & 48.3\% & 51.7\% \\
MNCL~\shortcite{liu2025mncl}    & 48.4\% & 79.0\% & 59.2\% & 40.8\% \\
\bottomrule
\end{tabular*}
}

\caption{Boundary-related evidence on the KITTI360Pose validation split.
(a) Prevalence of boundary-truncated instances and affected hints and queries.
(b) Exact Top-1 recall, Close@1 recall within $15\,\mathrm{m}$, and the
partition of Top-1 retrieval errors into close non-GT and other submaps.}
\label{tab:pipeline_diagnosis}
\end{table}
}

\newcommand{\tableCityLocTransfer}{
\begin{table}[t]
\centering
{\small
\setlength{\tabcolsep}{2pt}
\begin{tabular*}{\columnwidth}{@{\extracolsep{\fill}}ccccccc@{}}
\toprule
\multirow{2}{*}{Method}
& \multicolumn{3}{c}{CityLoc-K Val.} & \multicolumn{3}{c}{CityLoc-K Test} \\
\cmidrule(lr){2-4}\cmidrule(l){5-7}
& R@5m & R@10m & R@15m & R@5m & R@10m & R@15m \\
\midrule
Text2Pos & 16.48 & 40.69 & 62.92 & 14.62 & 38.27 & 59.55 \\
Text2Loc & 18.91 & 45.26 & 64.28 & 17.97 & 41.22 & 61.50 \\
MNCL & 19.30 & 45.94 & 64.50 & 18.76 & 42.63 & 62.58 \\
CMMLoc & 20.77 & 48.65 & 67.89 & 21.71 & 46.67 & 66.00 \\
VLM-Loc & \textbf{36.23} & \textbf{63.66} & \underline{77.77} & \textbf{35.91} & \textbf{63.81} & \underline{76.79} \\
\midrule
PARC-Loc & \underline{32.84} & \underline{61.12} & \textbf{80.47} & \underline{31.04} & \underline{59.31} & \textbf{77.64} \\
\bottomrule
\end{tabular*}
}
\caption{Localization on CityLoc-K. Recalls are reported as percentages under
$5\,\mathrm{m}$, $10\,\mathrm{m}$, and $15\,\mathrm{m}$ thresholds. Best results are in \textbf{bold}, and second-best
results are \underline{underlined}. VLM-Loc~\shortcite{kang2026vlmloc} is included for reference and uses
vision-language reasoning over BEV and scene-graph inputs, whereas the
remaining methods follow the conventional T2P setting.}
\label{tab:cityloc_k_transfer}
\end{table}
}

\newcommand{\tableCityLocCTransfer}{
\begin{table}[t]
\centering
{\small
\begin{tabular*}{\columnwidth}{@{\extracolsep{\fill}}cccc@{}}
\toprule
Method & R@5m $\uparrow$ & R@10m $\uparrow$ & R@15m $\uparrow$ \\
\midrule
Text2Pos & 8.11 & 27.21 & 50.01 \\
Text2Loc & 9.45 & 29.44 & 51.17 \\
MNCL & 13.68 & 35.93 & 53.78 \\
CMMLoc & 11.68 & 34.79 & 54.71 \\
VLM-Loc & \textbf{21.37} & \textbf{49.12} & \underline{68.26} \\
\midrule
PARC-Loc & \underline{16.56} & \underline{48.96} & \textbf{73.57} \\
\bottomrule
\end{tabular*}
}
\caption{Zero-shot cross-domain localization on CityLoc-C. Recalls are reported
as percentages. PARC-Loc is trained on CityLoc-K and evaluated on CityLoc-C. Best results are in \textbf{bold}, and second-best results are
\underline{underlined}.}
\label{tab:cityloc_c_transfer}
\end{table}
}

\title{PARC-Loc: Text-to-Point-Cloud Localization with Partial Assignment and Relational Consistency}
\author{Shengkai Ma \qquad Zhenyu Hou \qquad Weihua Cao}
\hypersetup{pdfauthor={Shengkai Ma, Zhenyu Hou, Weihua Cao}}

\date{}
\hypersetup{pdftitle={PARC-Loc: Text-to-Point-Cloud Localization with Partial Assignment and Relational Consistency},pdfsubject={Research preprint},pdfkeywords={text-to-point-cloud localization, spatial reasoning, partial assignment}}
\begin{document}
\twocolumn[\begin{@twocolumnfalse}
\begin{center}
{\LARGE\bfseries PARC-Loc: Text-to-Point-Cloud Localization with\\[3pt]
Partial Assignment and Relational Consistency\par}
\vspace{12pt}
{\large Shengkai Ma\qquad Zhenyu Hou\qquad Weihua Cao\par}
\end{center}
\begin{center}\begin{minipage}{0.94\textwidth}
\begin{abstract}
Text-to-point-cloud localization estimates a position in a city-scale 3D map from descriptions of surrounding objects.
Existing
coarse-to-fine methods retrieve submaps using aggregate learned compatibility
and then localize within a selected submap. However,
repetitive or similar urban objects can inflate the embedding similarity between the
query and multiple submaps, even when the instance layout within a submap violates the
query description. Meanwhile, query-relevant instances often span submap boundaries, leaving the
retrieved submap with incomplete contextual evidence.
We term these failure modes \emph{layout-inconsistent aliasing}
and \emph{boundary evidence incompleteness}, respectively.
To address them, we propose \textbf{PARC-Loc}, a coarse-to-fine localization
framework built on \emph{Partial Assignment with Relational Consistency}
(\textbf{PARC}). PARC jointly models hint-object compatibility and pairwise
spatial relations, allowing unmatched elements while favoring assignments
consistent with the queried layout. At the coarse stage, its candidate-level assessment complements neural similarity for layout-consistent submap selection. At the fine stage, the context is expanded with query-relevant instances from adjacent
submaps, while PARC yields object-level matching weights that guide
cross-modal attention.
Extensive experiments on KITTI360Pose and CityLoc show that PARC-Loc outperforms
conventional coarse-to-fine baselines. On KITTI360Pose, our method improves Top-1 localization recall at
$5\,\mathrm{m}$ from 0.50 to 0.67, achieving a 34\% relative gain over the strongest baseline.

\end{abstract}

\end{minipage}\end{center}
\vspace{1em}
\end{@twocolumnfalse}]
\section{Introduction}

Text-to-point-cloud (T2P) localization estimates a position in a city-scale 3D map from
linguistic descriptions of nearby landmarks~\cite{kolmet2022text2pos}. Unlike referred-object grounding, the
target does not necessarily correspond to an observed instance. For example, a query such as
``west of the green building and east of the gray road'' specifies the target through the joint
configuration of several surrounding objects. This provides a language-based interface for locating without exact addresses or GPS coordinates,
which is promising for applications in autonomous driving and embodied-agent \mbox{navigation~\cite{anderson2018vln,deruyttere2019talk2car}}. However, it remains challenging to realize such capabilities: grounding these descriptions in a large-scale
point-cloud map requires jointly aligning object semantics and spatial relations across language and point clouds~\cite{chen2022vil3dref}.

\begin{figure}[t]
\centering
\includegraphics[width=\linewidth]{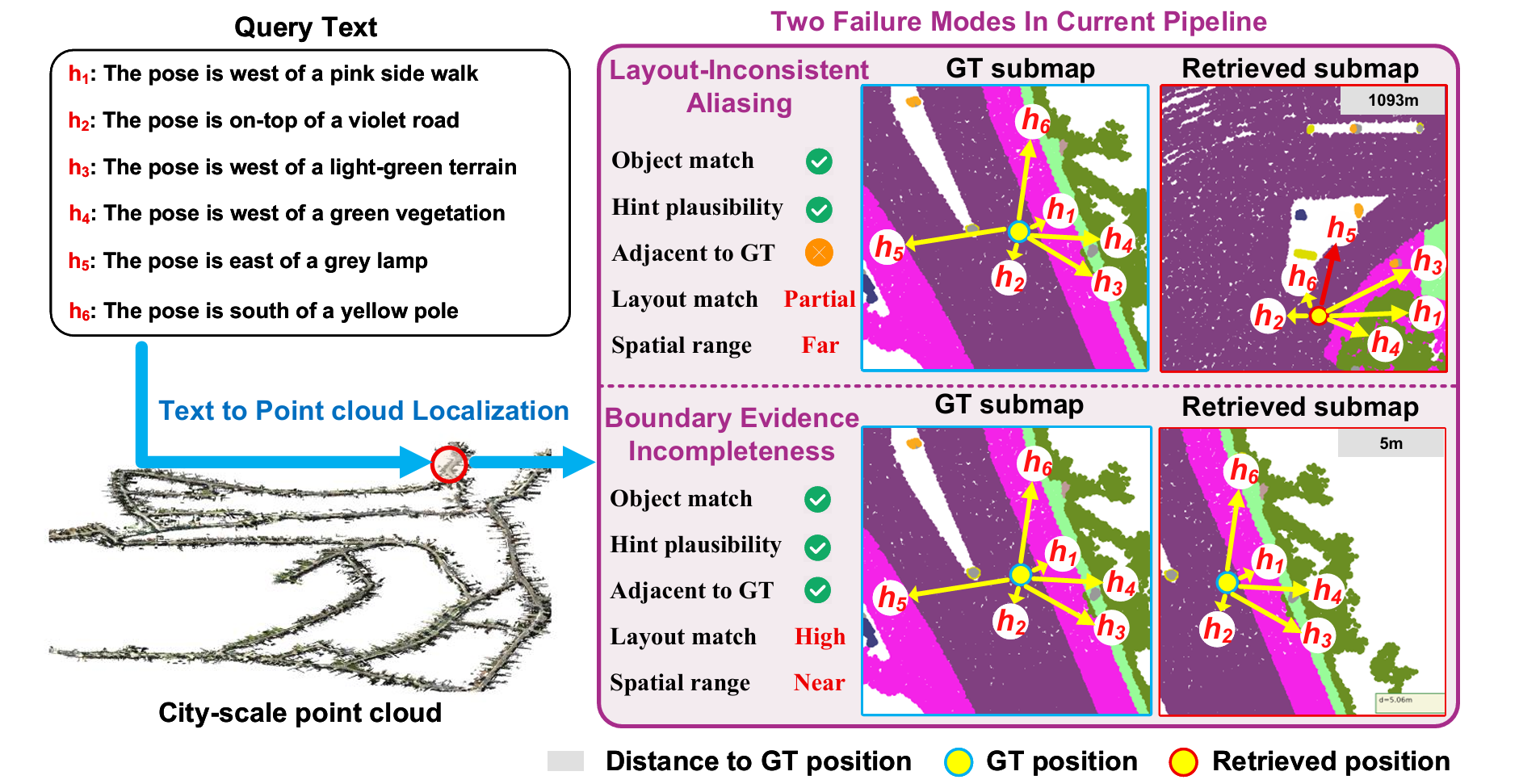}
\caption{Two failure modes in the standard coarse-to-fine decision structure.}
\label{fig:intro}
\end{figure}

To address this challenging task, most existing systems follow a coarse-to-fine design introduced by
Text2Pos~\cite{kolmet2022text2pos}: a neural retriever first selects candidate
submaps, after which a fine localizer predicts a position within a selected
submap. Subsequent methods advance this pipeline along several dimensions. For spatial structure encoding, RET explicitly models relations
at the hint and instance levels, while SpatiaLoc constructs spatial descriptors at instance and global levels~\cite{wang2023ret,shang2026spatialoc}.
For cross-modal representation alignment, Text2Loc adopts global text--submap contrastive learning, whereas MNCL extends
negative contrastive supervision to the global, instance, and relation levels~\cite{xia2024text2loc,liu2025mncl}.
For robustness to partial or ambiguous correspondence, CMMLoc captures semantic interactions under partial relevance, while PMSH propagates localization uncertainty across the coarse and fine stages~\cite{xu2025cmmloc,feng2025pmsh}.
Despite these advances, these methods generally retain the same decision making: coarse retrieval ranks candidates by an aggregate
query-submap similarity, while fine localization treats each selected submap as a self-contained evidence domain. Two limitations exist within this structure:
it implicitly equates aggregate similarity with a jointly valid layout and the selected submap with complete localization context.

The first limitation can lead to \emph{layout-inconsistent aliasing}. As illustrated
in Fig.~\ref{fig:intro}, common object categories and similar appearances across
locations can produce high aggregate query-submap similarity through individually
plausible object matches, even when their joint spatial configuration is inconsistent
with the relational layout described by the query. Although relation-aware encoders
incorporate spatial cues~\cite{wang2023ret,shang2026spatialoc}, the final score does not
explicitly verify relational consistency at the candidate level.
Specifically, these methods
do not check whether a query-to-instance configuration satisfies the relational
constraints in the query. This motivates an explicit assessment of
relational consistency over candidate-specific partial assignments during coarse selection.

The second limitation can lead to \emph{boundary evidence incompleteness}. Cell-based submap construction
may distribute query-relevant instances across neighboring
submaps, leaving any single submap with incomplete contextual evidence. This loss
of context can make neighboring candidates harder to distinguish and exacerbate the
text-submap matching uncertainty studied in prior work~\cite{xu2025cmmloc,feng2025pmsh}.
Incorporating instances from adjacent submaps can recover cross-boundary evidence. However, it may
also introduce distractors whose individual object matches appear plausible while their
relational configurations are inconsistent with the query. Fine localization should therefore
preserve the selected submap as the coordinate anchor while selectively incorporating
query-relevant context from adjacent submaps.

Although caused by different structural choices, the two failures create the same matching requirement: determining which subset of individually plausible hint--object correspondences forms a coherent explanation of the query layout. At the coarse stage, this assessment differentiates coherent candidates from aliases supported only by isolated coincidences; after neighboring-object aggregation makes cross-boundary evidence available, it separates useful context objects from the accompanying distractors. The required mechanism must therefore allow unsupported hints and irrelevant objects to remain unmatched while jointly evaluating unary attribute compatibility and pairwise layout consistency.
We therefore propose
PARC-Loc, a text-to-point-cloud localization framework built around \emph{Partial Assignment with
Relational Consistency} (PARC). PARC constructs a soft partial hint--object
assignment that jointly evaluates attribute compatibility and pairwise spatial
consistency while leaving missing or irrelevant elements unmatched.
This mechanism is applied to both coarse submap retrieval and fine localization.
In the coarse stage, PARC evaluates each neural
proposal and combines its explicit candidate-level consistency with the neural retrieval score
for submap selection. In the fine stage,
the selected submap defines the coordinate frame, while objects collected from
neighboring cells form an expanded evidence set. Reapplying PARC to this set
produces object-level matching weights that bias cross-attention toward
query-relevant context.

Taken together, these two stage-specific applications of PARC yield substantial gains.
On the KITTI360Pose test set, PARC-Loc improves Top-1@5m localization
recall from 0.50 to 0.67, a 34\% relative improvement over MNCL. We further conduct transfer and zero-shot experiments on CityLoc-K and CityLoc-C, where PARC-Loc achieves competitive performance.
Module-wise ablations support the individual contributions of
our method. In a controlled single-scene stress test,
the PARC selector retains 96.7\% of its Top-1 retrieval recall as the gallery
grows from 819 to 2,549 submaps, compared with 83.8--89.7\% for reported baselines.

Our main contributions are summarized as follows:
\begin{itemize}
    \item We thoroughly analyze the standard coarse-to-fine
    pipeline adopted in T2P localization and diagnose two structural failure modes: \emph{layout-inconsistent aliasing} during coarse submap retrieval and
    \emph{boundary evidence incompleteness} during fine localization.

    \item We propose PARC-Loc, a novel language-based localization framework.
    Within this framework, \emph{Layout-Consistent Submap Selection} and \emph{Cross-Boundary Context Localization} specifically mitigate the two structural failure modes through
    \emph{Partial Assignment with Relational Consistency} (PARC), a soft
    hint--object assignment that jointly evaluates semantic and spatial compatibility under partial correspondence.

    \item Extensive experiments on the KITTI360Pose dataset demonstrate that our method outperforms state-of-the-art methods.
    Experiments on the CityLoc-K and CityLoc-C datasets and under controlled gallery growth further assess the transfer and robustness of our method.
\end{itemize}

\section{Related Work}

\paragraph{Text-to-point-cloud localization.}
Text2Pos establishes the prevailing coarse-to-fine formulation on
KITTI360Pose, retrieving candidate submaps before estimating a target offset
from hint--instance associations~\cite{kolmet2022text2pos}. Subsequent methods
address different sources of cross-modal ambiguity: RET models instance
relations, Text2Loc adopts stronger language representations and matching-free
localization, and MNCL introduces multi-level negative contrastive learning
over global, instance, and relation cues~\cite{wang2023ret,xia2024text2loc,liu2025mncl}.
CMMLoc and PMSH model partial correspondence and
uncertainty~\cite{xu2025cmmloc,feng2025pmsh}, while SpatiaLoc and SympLoc
strengthen spatial descriptors and alignment
geometry~\cite{shang2026spatialoc,shang2026symploc}. Vehicle-SceneInteraction
studies text-driven 3D LiDAR place recognition for autonomous
driving~\cite{shang2026vehicle}, whereas VLM-Loc performs localization through
BEV maps and scene graphs~\cite{kang2026vlmloc}. These methods improve the
representation, training, or retrieval of submap compatibility. PARC-Loc
instead validates joint layouts during coarse selection and prevents the
selected submap boundary from limiting fine-stage evidence.

\paragraph{Language-guided 3D spatial reasoning.}
Language-guided 3D grounding identifies objects or regions from textual
descriptions~\cite{huang2022mvt,jain2022butddetr}. ScanRefer and ReferIt3D
establish indoor benchmarks~\cite{chen2020scanrefer,achlioptas2020referit3d},
followed by methods for relation modeling, spatial--language alignment,
progressive point selection, and dense
alignment~\cite{he2021transrefer3d,zhao20213dvg,roh2022languagerefer,luo20223dsps,wu2023eda}.
CityRefer extends grounding to city-scale point clouds~\cite{miyanishi2023cityrefer}.
Unlike referred-object grounding, the target in T2P localization may not
correspond to an observed instance; it must be inferred from the joint
configuration of surrounding objects. This distinction motivates validating
hint--object correspondences jointly rather than grounding each hint independently.

\section{Preliminaries and Pipeline Diagnosis}
\label{sec:residual_analysis}

We formulate text-to-point-cloud localization and diagnose two limitations of
its standard coarse-to-fine pipeline.

\subsection{Text-to-Point-Cloud Localization}
Let $M=\{M_n\}_{n=1}^{N}$ denote a city-scale point-cloud map partitioned into
overlapping submaps, where
$M_n=\{P_{n,j}\}_{j=1}^{O_n}$ contains $O_n$ segmented instances. Given a query
$q=\{h_i\}_{i=1}^{H}$, whose hints describe target-to-object relations, the
task is to estimate the target position $(x,y)$ from the joint configuration
of the referenced objects. Most existing methods first rank submaps by
query--submap similarity and then predict the position from $q$ and the
instances in the selected submap $M_{\hat n}$
~\cite{kolmet2022text2pos,xia2024text2loc,liu2025mncl,xu2025cmmloc}.

Despite its practical effectiveness, this pipeline leaves two structural gaps:
(1) a submap is retrieved through an aggregate embedding comparison without explicitly verifying the uniqueness and full consistency between the query and the instance layout.
(2) fine localization infers the target position within a single submap, ignoring that the context is often incomplete because of each submap's fixed spatial extent.

\begin{figure}[t]
\centering
\includegraphics[width=\linewidth]{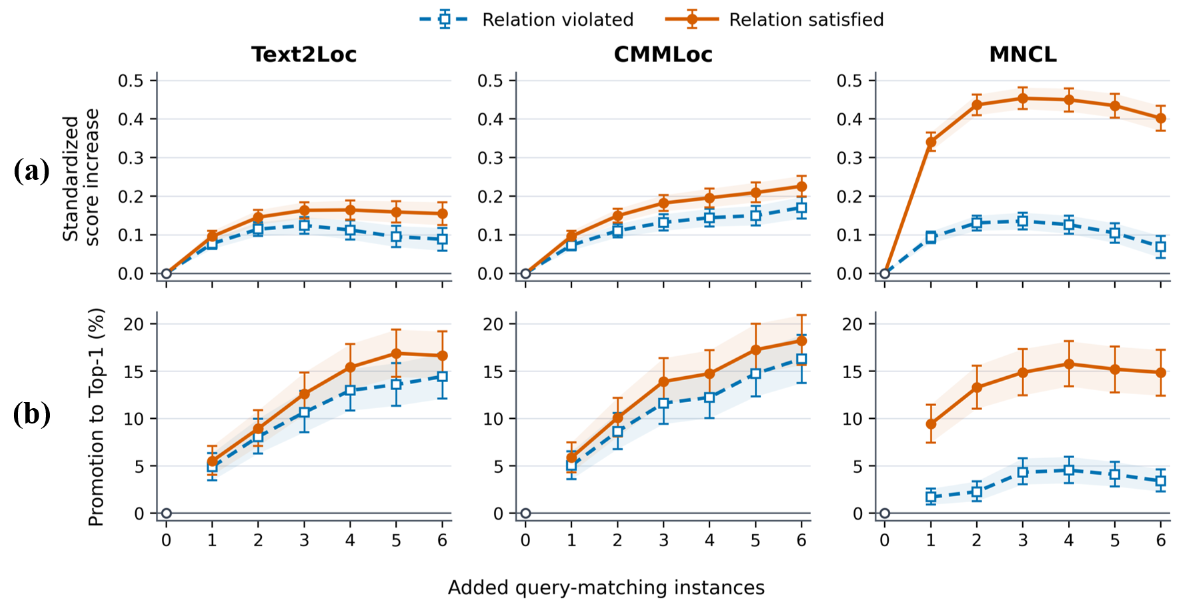}
\caption{Effect of added query-matching instances on candidate scores. Top:
standardized score change; bottom: promotion rate among initially non-Top-1
candidates. Curves compare relation-satisfied and relation-violated
placements; error bars denote 95\% cluster-bootstrap confidence intervals.}
\label{fig:diagnosis1}
\end{figure}
\tablePipelineDiagnosis

\subsection{Diagnosis of Pipeline Limitations}
We analyze the two gaps using released Text2Loc, CMMLoc, and MNCL models~\cite{xia2024text2loc,liu2025mncl,xu2025cmmloc}. All analyses are conducted with their official checkpoints on the KITTI360Pose dataset~\cite{kolmet2022text2pos}.

\paragraph{Partial Matches Elevate Inconsistent Candidates.}
Urban environments typically contain abundant similar or repetitive objects~\cite{torii2013repetitive}.
Repeated objects allow a candidate to contain individually plausible
but jointly inconsistent matches, which can raise the aggregate
submap score when no explicit joint-layout verification is imposed.
To measure this effect, we test 911 partially matched candidates by
replacing unrelated instances with one to six exact semantic matches. Paired
variants have identical geometry and semantic attributes but differ in whether their
placement satisfies the queried relation. Figure~\ref{fig:diagnosis1} shows
that both candidate scores and Top-1 promotion rates increase even under
relation-violated placements, while satisfying the relation provides an
additional gain. These results demonstrate that partial matches may mislead models without explicit joint layout verification, resulting in
\emph{layout-inconsistent aliasing}.

\paragraph{Submap Boundaries Truncate Query Evidence.}
Large-scale objects such as sidewalks and roads often span multiple submaps
and are therefore truncated at cell boundaries.
Table~\ref{tab:pipeline_diagnosis}(a) shows that
40.28\% of instances are boundary-truncated, 64.49\% of hints refer to them,
and 99.75\% of queries contain at least one such hint.
The resulting fragments make
neighboring submaps similar while reducing the evidence available within any
single submap. Across the three
baselines, nearby non-ground-truth submaps account for 46.8--59.2\% of Top-1
retrieval errors (Table~\ref{tab:pipeline_diagnosis}(b)), indicating that
neighboring candidates often share relevant evidence while omitting different
fragments.
Thus, even when the coarse stage retrieves the submap containing
the most relevant evidence, the fine stage may still operate on incomplete
context, resulting in
\emph{boundary evidence incompleteness}.

Together, these diagnoses define two requirements: coarse selection should
validate matched instances as a joint layout, whereas fine localization should
preserve the selected coordinate frame while admitting relevant
cross-boundary evidence. Both require semantic and spatial compatibility to be
evaluated under partial correspondence. This motivates PARC's candidate-level
assessment for submap selection and object-level weights for context routing.

\section{Method}
\label{sec:method}

\begin{figure*}[t]
\centering
\includegraphics[width=\textwidth]{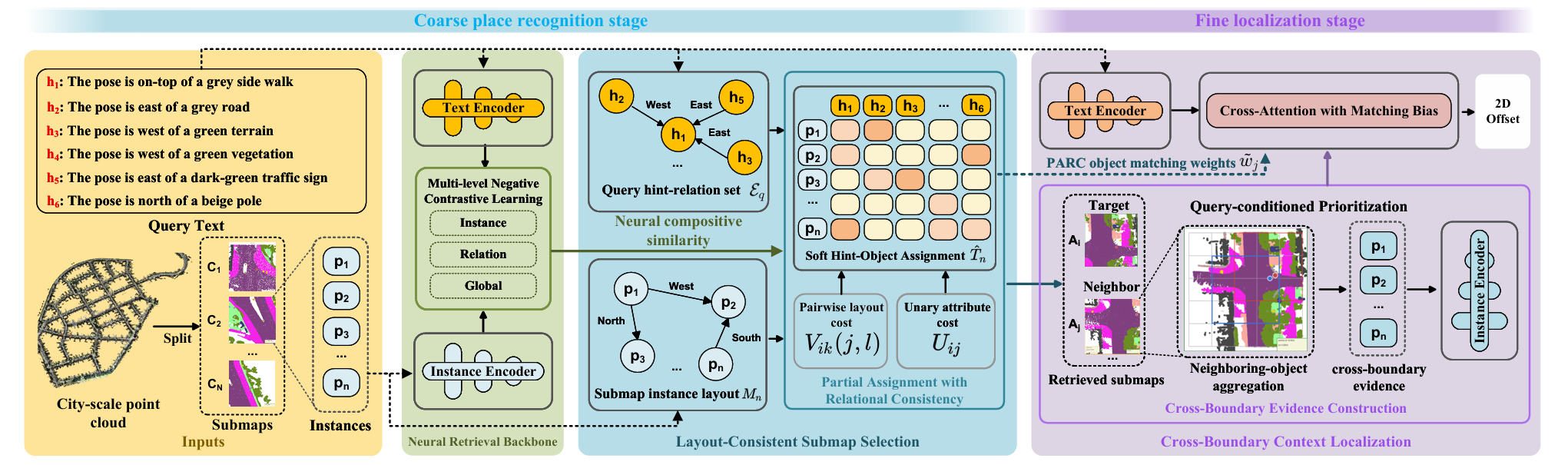}
\caption{Overview of PARC-Loc. In the coarse stage, the relational consistency between hints and each submap is explicitly assessed through partial assignment to produce a reliable
submap selection. In the fine stage, neighboring-object aggregation first makes cross-boundary context available; PARC
is then reapplied to produce object weights that bias cross-attention before offset prediction.}
\label{fig:framework}
\end{figure*}

\subsection{Overview}

Given a query $q$, the coarse retrieval stage first uses a neural retriever to rank
the submaps in $M$ by neural similarity scores $S_{\mathrm{sim}}(q,M_n)$ and retains the top-$K$ candidates. For each candidate $M_n$,
PARC evaluates whether the objects form a partial hint--object assignment that is explicitly consistent in both
attributes and pairwise layout. Combining its candidate-level score with neural similarity, our approach identifies the best-matching submap $M_{\hat n}$.

At the fine localization stage, the selected submap \(M_{\hat n}\) serves as the
coordinate anchor. Objects from its neighborhood are expressed in its coordinate
frame, and PARC is reapplied to compute query-conditioned object-matching weights.
These weights bias hint-to-object cross-attention before the model predicts a location offset.

\subsection{Partial Assignment with Relational Consistency (PARC)}
\label{sec:parc}

At both stages, a query refers to only a subset of map objects, while some
referenced objects may be absent or truncated and unrelated objects may be
present. PARC models this partial correspondence while requiring the assigned
objects to preserve the relations expressed by the query. It combines unary
attribute compatibility, pairwise layout consistency~\cite{memoli2011gromov,peyre2016gromov}, and partial assignment
constraints in one objective.

Consider a query $q$ with $H$ hints and an input instance set
$P=\{P_j\}_{j=1}^{O}$. Hint $i$
specifies a category $\ell_i$, an appearance attribute $\kappa_i$, and a
target-to-object direction. Instance $P_j$ has category $\ell_j$, appearance
$\kappa_j$, and a two-dimensional ground-plane center $p_j$. We represent their
soft correspondence by a nonnegative matrix $T\in\mathbb{R}^{H\times O}$, where $T_{ij}$ is
the assignment mass between hint $i$ and instance $P_j$.

\paragraph{Unary attribute compatibility.}
Let $\delta(x,y)$ be zero if $x=y$ and one otherwise. The unary cost of an
individual correspondence is
\begin{equation}
U_{ij}
=\lambda_{\mathrm{cat}}\delta(\ell_i,\ell_j)
+\lambda_{\mathrm{app}}\delta(\kappa_i,\kappa_j),
\label{eq:matching_unary}
\end{equation}
where $\lambda_{\mathrm{cat}}$ and $\lambda_{\mathrm{app}}$ specify mismatch
cost weights. Unary compatibility identifies plausible pairs but, as diagnosed in
Sec.~\ref{sec:residual_analysis}, cannot determine whether they form one valid
layout.

\paragraph{Pairwise layout consistency.}
We therefore evaluate the relations among assigned objects. From the
target-to-object directions in $q$, we derive a set $\mathcal{E}_q$ of valid
hint pairs. Each $(i,k)\in\mathcal{E}_q$ implies an object-to-object relation
$r^q_{ik}$. For example, if the
target is west of one referenced object and east of another, the first object
should lie east of the second. For $(i,k)\in\mathcal{E}_q$, the cost
$V_{ik}(j,l)$ penalizes assigning the hint pair $(i,k)$ to the object pair
$(j,l)$ when the ground-plane relation between instances $P_j$ and $P_l$ is
inconsistent with the object-to-object relation $r^q_{ik}$ implied by hints
$h_i$ and $h_k$.
Following the fused transport principle
of combining node attributes and pairwise structure~\cite{vayer2019optimal},
we define
\begin{equation}
\mathcal{J}_{q,P}(T)
=\sum_{i,j}U_{ij}T_{ij}
+\frac{\alpha_{\mathrm{rel}}}{Z_q}
\sum_{(i,k)\in\mathcal{E}_q}\sum_{j,l}
V_{ik}(j,l)T_{ij}T_{kl},
\label{eq:matching_objective}
\end{equation}
where $Z_q=\max(1,|\mathcal{E}_q|)$ normalizes the number of valid query relations and
$\alpha_{\mathrm{rel}}$ controls the relative penalty for layout violations.

\paragraph{Partial correspondence.}
A query mentions only a subset of map objects, and boundary truncation may
remove referenced objects. A complete one-to-one assignment would force missing hints or unrelated
objects into incorrect correspondences. We therefore impose partial rather than complete
matching~\cite{chapel2020partialot,sarlin2020superglue}:
\begin{equation}
\left\{
\begin{aligned}
T_{ij}&\geq 0,\quad \sum_jT_{ij}\leq 1,\quad \sum_iT_{ij}\leq 1,\\
\sum_{i,j}T_{ij}&=\mu,\qquad \mu=\min(\rho H,O).
\end{aligned}
\right.
\label{eq:matching_constraints}
\end{equation}
where $\rho\in[0,1]$ specifies the matched fraction of hint mass. The
capacities limit repeated use of a hint or object, while the partial mass
$\mu$ leaves irrelevant objects and missing references unmatched.
We define
$\hat T(q,P)=\operatorname{PARC}(q,P)$ as a feasible plan obtained by approximately minimizing the objective in Eq.~\eqref{eq:matching_objective} subject to the constraints in Eq.~\eqref{eq:matching_constraints}.
Its normalized objective value
provides the candidate-level consistency used in the coarse stage, while its column mass
provides the object matching weights used in fine localization.

\subsection{Coarse Stage: Layout-Consistent Submap Selection}

Neural retrieval efficiently narrows the city-scale search space~\cite{uy2018pointnetvlad,komorowski2021minkloc3d}, while PARC explicitly
validates whether the matched objects form a coherent query layout in each of the
$K$ candidate submaps. For each candidate $M_n$, PARC returns
$\hat T_n=\operatorname{PARC}(q,M_n)$. Its normalized score is
\begin{equation}
S_{\mathrm{parc}}(q,M_n)
=-\frac{\mathcal{J}_{q,M_n}(\hat T_n)}
{\max\!\left(\lVert\hat T_n\rVert_1,\varepsilon_0\right)},
\label{eq:matching_score}
\end{equation}
where $\varepsilon_0>0$ is a numerical stabilizer. A high score requires low-cost unary matches that also preserve the joint
layout. Because PARC complements rather than replaces learned neural similarity, we combine its candidate-level consistency $S_{\mathrm{parc}}$ with the neural similarity $S_{\mathrm{sim}}$ to select the submap. This design effectively suppresses layout-inconsistent aliases. The selected submap defines the coordinate frame for fine localization, but not its complete evidence domain.

\subsection{Fine Stage: Cross-Boundary Context Localization}

Even a correct or nearby selected submap may omit query-relevant instances
across its boundary. The fine stage therefore first expands the available
context and then uses PARC to determine which objects support the joint query
layout before routing them through cross-attention.

\paragraph{Neighboring-object aggregation.}
We collect objects from the neighborhood centered at $M_{\hat n}$ and
express every object in the selected-submap frame:
\begin{equation}
\tilde p_j
=\frac{p_j-o_{\hat n}}{L_{\hat n}},
\label{eq:selected_submap_coordinates}
\end{equation}
where $o_{\hat n}$ is the two-dimensional origin and $L_{\hat n}$ is the side length of
$M_{\hat n}$. This preserves a single coordinate system across submap boundaries. We retain
the most relevant objects using a prioritization strategy that considers their
compatibility with the query and the surrounding spatial context. This deterministic capacity
control makes cross-boundary evidence available without asserting that every
retained object is relevant.

\paragraph{Object matching weights.}
Because the expanded context also introduces irrelevant objects, we reapply
PARC to the retained instance set $P_{\mathrm{ctx}}$, obtaining
$\hat T_{\mathrm{ctx}}=\operatorname{PARC}(q,P_{\mathrm{ctx}})$. Its
object-wise column mass and normalized weight are
\begin{equation}
w_j=\sum_i(\hat T_{\mathrm{ctx}})_{ij},
\qquad
\tilde w_j=\frac{w_j}{\max_l w_l+\varepsilon_0}.
\label{eq:object_matching_weight}
\end{equation}
The normalized weight quantifies the contribution of instance $P_j$ to a
low-cost joint explanation.
These quantities are matching weights, which enable the fine localizer to route
the available context toward query-relevant objects.

\paragraph{Cross-attention with matching bias.}
Let $q_i$ and $k_j$ denote the $d_k$-dimensional projected representation of
hint $h_i$ and the key projection of context instance $P_j$, respectively. We incorporate the PARC-derived weight as an additive log bias
in hint-to-object attention:
\begin{equation}
A_{ij}
=
\operatorname{softmax}_{j}\!\left(
\frac{q_i^\top k_j}{\sqrt{d_k}}
+
\beta\log(\tilde w_j+\varepsilon_0)
\right),
\label{eq:matching_biased_attention}
\end{equation}
where $\beta\geq 0$ controls the strength of the matching bias. When
$\beta=0$, Eq.~\eqref{eq:matching_biased_attention}
reduces to standard content-based attention~\cite{vaswani2017attention}. Increasing $\beta$ gives more
attention to objects supported by a relationally consistent partial
assignment while retaining the flexibility of learned cross-modal
interaction.

The fine decoder produces a normalized two-dimensional offset
$\hat d$ in the selected-submap frame, which is
converted back to map coordinates by
\begin{equation}
\hat p=o_{\hat n}+L_{\hat n}\hat d.
\label{eq:position_recovery}
\end{equation}
The neural retriever retains its standard training objective, while the fine
decoder is trained with the fine-localization loss
$\mathcal{L}_{\mathrm{fine}}=\lVert\hat d-d^{*}\rVert_2^2$,
where $d^{*}=(p^*-o_{\hat n})/L_{\hat n}$ is the
ground-truth normalized offset and $p^*$ is the ground-truth map
position. At inference,
PARC first refines the neural
candidate ranking and is then reapplied to route cross-boundary evidence
during offset prediction.

\section{Experiments}
\label{sec:experiments}
\tableFullLocalization
\tableCoarseRetrieval
\subsection{Experimental Setup}

\paragraph{Dataset and protocol.}
Experiments are conducted on KITTI360Pose~\cite{kolmet2022text2pos,liao2022kitti360} and
CityLoc~\cite{kang2026vlmloc}, two benchmarks for city-scale
T2P localization. Following the official KITTI360Pose protocol,
we use five scenes for training, one for validation, and three for testing,
with $30\,\mathrm{m}$ cubic submaps extracted at a $10\,\mathrm{m}$ stride. CityLoc comprises CityLoc-K for training and
in-domain evaluation and CityLoc-C~\cite{hu2022sensaturban} for zero-shot cross-domain testing. Each
$50\times50\,\mathrm{m}$ map is paired with six textual hints.
KITTI360Pose contains 3,187 validation queries and 11,505 test queries,
corresponding to 1,116 and 3,932 distinct target-position clusters, respectively.

\paragraph{Evaluation metrics.}
Submap retrieval is evaluated by top-1/3/5 recall of the annotated
ground-truth submap. Fine localization uses the top-$k$ candidates,
$k\in\{1,5,10\}$, and reports recall within
$\tau\in\{5,10,15\}\,\mathrm{m}$~\cite{kolmet2022text2pos,liu2025mncl}.

\paragraph{Implementation.}
PARC-Loc adopts the neural backbone of MNCL~\cite{liu2025mncl}.
For hyper-parameters, we set $\alpha_{\mathrm{rel}}=0.8$,
$\lambda_{\mathrm{cat}}=1$, and $\lambda_{\mathrm{app}}=0.5$.
The benchmark experiments were run on an NVIDIA RTX A6000 GPU; the
matched-seed analysis below used an NVIDIA A800 80GB. We make no runtime
comparison across GPU types.

For KITTI360Pose, neural retrieval proposes 20 candidates for PARC reranking.
The fine decoder uses at most 64 object tokens from the selected-submap-centered
$3\!\times\!3$ neighborhood. Exact category--appearance matches are retained
first, category matches second, and remaining objects third, with deterministic
source order and distance used to break ties. This ordering controls the token
budget; PARC subsequently computes matching weights over all retained objects.

\subsection{Quantitative Analysis}
\paragraph{End-to-end localization.}
As shown in Table~\ref{tab:main_results}, PARC-Loc obtains the best
recall across all metrics on both KITTI360Pose splits. Under the strict
Top-1@$5\,\mathrm{m}$ metric, it reaches 0.71 on validation and 0.67 on test,
exceeding MNCL by 17 percentage points on both splits. Its advantage persists across larger thresholds and candidate
ranks, although the margin narrows as recall approaches saturation. These results demonstrate the overall effectiveness of our method. The analyses below further disentangle the sources of the observed gains.

\paragraph{Submap retrieval.}
Table~\ref{tab:coarse_retrieval} shows that PARC-Loc outperforms the strongest baseline
across all retrieval metrics, by 4--5 percentage points on validation and
6--7 percentage points on test. By
construction, this metric credits only
the annotated ground-truth submap. A neighboring non-GT submap can still
support a correct final prediction once cross-boundary context is included.
This retrieval
metric is used to isolate coarse-stage submap selection, while final
localization remains the primary measure.

\subsection{Ablation Studies}
We evaluate the PARC terms and assess the
complete pipeline by cumulatively enabling its coarse and fine stage
components.
\paragraph{Ablation on Submap Retrieval.}
\tableCoarseAblation
All variants in Table~\ref{tab:coarse_ablation} start from the same neural
backbone and differ only in the objective used in PARC. Full PARC raises Top-1 retrieval recall from 0.49 to 0.54 on validation and
from 0.46 to 0.51 on test. Replacing partial
assignment causes the largest test degradation, reducing Top-1 recall to 0.48.
Removing relational consistency lowers Top-1 recall to 0.50, while removing unary compatibility
ties full PARC at Top-1 but reduces Top-5 recall from 0.83 to 0.82.
The degradation of the relation and
partial variants further supports assessing a relationally consistent partial
layout rather than relying only on individually compatible objects. These
results demonstrate that explicit assessment of relational consistency through
partial assignment effectively constrains the joint layout of candidates,
achieving considerable improvements in submap retrieval.

\paragraph{Ablation on End-to-End Localization.}
\tableFineAblation
The first row of Table~\ref{tab:fine_ablation} is the controlled neural
baseline, which combines the MNCL-style neural retriever with the original
single-submap regressor. Coarse PARC raises test
Top-1@$5\,\mathrm{m}$ recall from 0.51 to 0.61. With coarse PARC fixed,
neighboring context, query-conditioned prioritization, and matching bias
increase it successively to 0.64, 0.66, and 0.67. These order-conditioned
increments show that the coarse-stage PARC provides the largest gain, while context
construction and routing contribute a further 6 percentage points. The two tables therefore
show the effectiveness of the shared PARC formulation and its two stage-specific readouts:
candidate-level energy for coarse assessment and object-level assignment mass
for fine-stage routing.

\paragraph{Context availability.}
To examine how much evidence is lost at a submap boundary, we compare three
context constructions on a frozen validation candidate stream and checkpoint
(Table~\ref{tab:context_availability}). This diagnostic complements the cumulative
pipeline ablation above; its rows are not intended as a row-wise reconstruction
of Table~\ref{tab:fine_ablation}.

\begin{table}[t]
\centering
\caption{Validation Top-1 localization recall (\%) under controlled context
construction. Each column gives the distance threshold; higher is better.}
\label{tab:context_availability}
{\small
\setlength{\tabcolsep}{3pt}
\begin{tabular*}{\columnwidth}{@{\extracolsep{\fill}}lrrr@{}}
\toprule
Context construction & @5m & @10m & @15m \\
\midrule
Single selected submap & 62.22 & 82.65 & 86.41 \\
Five-cell context & \textbf{67.74} & \textbf{84.00} & \textbf{86.57} \\
Distance-prioritized $3\!\times\!3$ & 66.68 & 83.68 & 86.32 \\
\bottomrule
\end{tabular*}}
\end{table}

Both expanded contexts improve the strict 5m endpoint over the single-submap
input, by 5.52 and 4.46 percentage points, respectively. However, the larger
distance-prioritized neighborhood does not outperform the five-cell context.
The comparison therefore supports making neighboring evidence available,
without implying that a larger context alone is always better.

\paragraph{Stability across training seeds.}
We next assess whether the matching-bias benefit persists across training
randomness. Table~\ref{tab:seed_stability} reports matched epoch-3 runs with
seeds 2027, 2028, and 2029. The $\beta=0.3$ condition has higher mean recall
at every reported candidate rank. With only three seeds, these results provide
a limited stability check, not a population-level significance claim.

\begin{table}[t]
\centering
\caption{Matched-seed localization recall within 5m (\%). Values are
mean $\pm$ sample standard deviation over seeds 2027--2029 at epoch 3.
Higher is better.}
\label{tab:seed_stability}
{\small
\begin{tabular*}{\columnwidth}{@{\extracolsep{\fill}}lrr@{}}
\toprule
Candidate rank & $\beta=0$ & $\beta=0.3$ \\
\midrule
Top-1 & $68.92\pm0.73$ & $71.10\pm0.49$ \\
Top-3 & $84.96\pm0.51$ & $85.90\pm0.79$ \\
Top-5 & $89.11\pm0.54$ & $90.97\pm0.68$ \\
Top-10 & $92.15\pm0.46$ & $93.95\pm0.50$ \\
\bottomrule
\end{tabular*}}
\end{table}

\begin{figure*}[t]
\centering
\includegraphics[width=\textwidth]{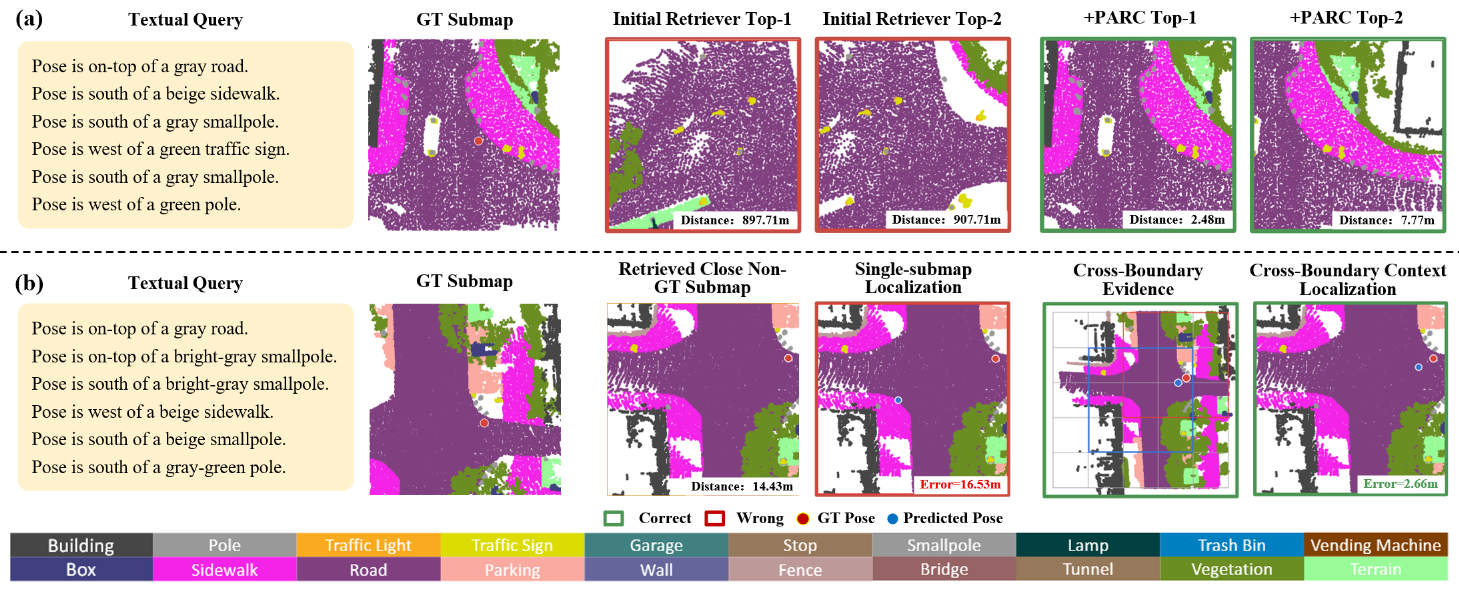}
\caption{Qualitative examples of the two stage-specific mechanisms.
Query boxes reproduce the relations while omitting the repeated subject
``Pose is.'' (a) PARC replaces the MNCL neural Top-1 distant alias with a
candidate within 15m. (b) A close non-GT selected submap omits neighboring
evidence; the $3\!\times\!3$ context reduces localization error. Red and green
outlines mark incorrect and correct results, respectively. The yellow marker
denotes the ground-truth pose, and the blue marker denotes the predicted pose.}
\label{fig:qualitative}
\end{figure*}

\subsection{Further Analysis}
\paragraph{Generalization on CityLoc.}
\tableCityLocTransfer
\tableCityLocCTransfer
Table~\ref{tab:cityloc_k_transfer} reports in-domain validation and test
results on CityLoc-K. PARC-Loc obtains the best R@$15\,\mathrm{m}$, reaching 80.47\%
on validation and 77.64\% on test, while VLM-Loc remains stronger at the
stricter R@$5\,\mathrm{m}$ and R@$10\,\mathrm{m}$ thresholds.

We further evaluate zero-shot cross-domain transfer from CityLoc-K to CityLoc-C without training or threshold tuning on CityLoc-C. As shown in
Table~\ref{tab:cityloc_c_transfer}, PARC-Loc obtains the best R@15m at 73.57\%
and remains within 0.16 percentage points of VLM-Loc at R@10m, while VLM-Loc
leads under the strict R@5m criterion. Given that VLM-Loc is tailored to vision-language
reasoning over BEV and scene-graph inputs~\cite{kang2026vlmloc}, this competitive performance supports the cross-domain
transferability of PARC-Loc without implying uniform superiority across all localization thresholds.

\paragraph{Robustness to gallery growth.}
\begin{figure}[t]
\centering
\includegraphics[width=\linewidth]{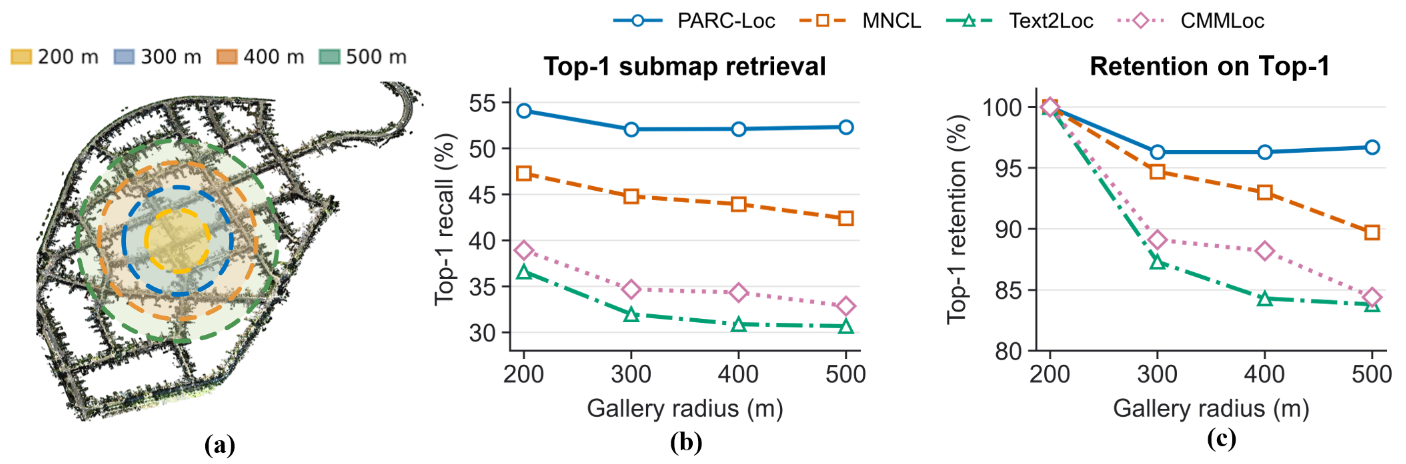}
\caption{Robustness to controlled gallery growth. (a) Gallery expansion. (b) Top-1 submap retrieval recall at each radius. (c) Recall retention relative to the \(200\,\mathrm{m}\) gallery.}
\label{fig:scalability}
\end{figure}
Following MNCL~\cite{liu2025mncl}, we fix 2,373 query descriptions within the \(200\,\mathrm{m}\) region and progressively expand the gallery around the same center. As shown in Fig.~\ref{fig:scalability}, PARC-Loc retains 96.7\% of its Top-1 recall at \(500\,\mathrm{m}\), decreasing from 54.11\% to 52.34\%. In comparison, MNCL, Text2Loc, and CMMLoc retain only 89.7\%, 83.8\%, and 84.4\%, respectively. This smaller degradation indicates that explicit layout validation improves robustness to distractors arising from repeated instances and partial layout matches.

\subsection{Qualitative Analysis}
\label{sec:qualitative}
The quantitative analyses measure aggregate performance; we now examine two
fixed cases to illustrate the two stage-specific mechanisms of PARC-Loc
(Figure~\ref{fig:qualitative}). In the first case, the neural Top-1 candidate
is a distant layout-inconsistent alias, and PARC promotes a candidate within
15m of the target. In the second, a nearby selected submap is not the annotated
ground-truth cell and omits neighboring evidence; cross-boundary context
reduces the final localization error. Together, these cases illustrate why
candidate verification and context expansion address different failure modes.
They are case-level evidence, not an estimate of how frequently either
mechanism corrects an error.

\section{Conclusion}

We presented PARC-Loc to address layout-inconsistent aliasing and boundary evidence
incompleteness in coarse-to-fine text-to-point-cloud localization. PARC evaluates attribute
and relational compatibility under partial correspondence: its candidate-level assessment complements
neural similarity for coarse selection, while reapplying PARC to cross-boundary context yields object-level weights
for localization in the selected-submap frame. PARC-Loc achieves 0.67 Top-1@$5\mathrm{m}$ test recall on KITTI360Pose, while ablations, CityLoc transfer, and gallery-growth tests support its effectiveness, transferability, and distractor robustness.

\clearpage
\bibliographystyle{plainnat}
\bibliography{references}
\end{document}